\documentclass[10pt,twocolumn]{article}

\usepackage[a4paper,margin=0.72in,columnsep=0.25in]{geometry}
\usepackage{times}
\usepackage{xeCJK}
\setCJKmonofont{FandolFang-Regular.otf}
\usepackage{microtype}
\usepackage{booktabs}
\usepackage{array}
\usepackage{multirow}
\usepackage{amsmath}
\usepackage{graphicx}
\usepackage{xcolor}
\usepackage{hyperref}
\usepackage{enumitem}
\usepackage{caption}
\usepackage{subcaption}
\usepackage{pgfplots}
\usepackage{tikz}
\usetikzlibrary{arrows.meta,positioning,shapes.misc}
\pgfplotsset{compat=1.18}

\definecolor{navy}{HTML}{183153}
\definecolor{blue}{HTML}{2F6FED}
\definecolor{cyan}{HTML}{28A7A1}
\definecolor{orange}{HTML}{E8833A}
\definecolor{errorred}{HTML}{C62828}
\definecolor{light}{HTML}{F2F6FB}
\newcommand{\miss}[1]{\textcolor{errorred}{\textbf{\texttt{#1}}}}
\newcommand{\hit}[1]{\textcolor{black}{\textbf{\texttt{#1}}}}
\newcommand{\zhmiss}[1]{\textcolor{errorred}{\texttt{#1}}}
\newcommand{\zhhit}[1]{\textcolor{black}{\texttt{#1}}}
\hypersetup{colorlinks=true,linkcolor=navy,citecolor=blue,urlcolor=blue}
\setlist{nosep,leftmargin=*}
\title{\textbf{KeyASRBench: Measuring Whole-Phrase Keyword\\Preservation in Real-World ASR}}
\author{%
Guanzheng Wei$^{1,\dagger}$ \quad Ziyi Zhao$^{2,\dagger}$\\
{\small $^1$Southwest University \quad $^2$University of Chinese Academy of Social Sciences}\\
{\small \texttt{wgz021228@email.swu.edu.cn} \quad \texttt{s20241040534@ucass.edu.cn}}\\
{\small $^\dagger$These authors contributed equally.}
}
\date{}

\begin{document}

\twocolumn[
  \begin{@twocolumnfalse}
    \maketitle
    \begin{abstract}
    Public automatic speech recognition (ASR) benchmarks increasingly report low sequence error rates, yet production systems still use hotword lists, contextual biasing, retrieval, and post-correction. This motivates a complementary question: does a transcript preserve selected names, products, technical terms, versions, and relational phrases? We present \textbf{KeyASRBench}, a bilingual pilot that operationalizes occurrence-based Keyword Error Rate (KER) over frozen whole-phrase annotations. The 10-hour model-conditioned challenge set contains 1,501 Chinese and English clips with 4,723 keyword occurrences. Across five locally inferred checkpoints and two official API systems, language-specific MER/WER ranges from 5.31\% to 12.91\%, while KER ranges from 13.61\% to 33.81\%. A disjoint 30-minute model-blind control yields KER of 10.65\%--27.31\%. The control confirms that selected-phrase errors extend beyond challenge-set selection.
    \end{abstract}
    \vspace{0.6em}
  \end{@twocolumnfalse}
]

\section{Introduction}

Large-scale weak supervision, speech foundation models, and speech-language models have substantially advanced general-purpose ASR. Whisper was trained on 680,000 hours of multilingual and multitask supervision and demonstrated strong zero-shot transfer across standard datasets~\cite{radford2023whisper}. More recent systems unify language identification, streaming and offline recognition, long-form transcription, and dialect coverage. Qwen3-ASR supports 52 languages and dialects and reports low error rates on a broad suite of public benchmarks~\cite{shi2026qwen3asr}. GLM-ASR-Nano-2512 is a compact 1.5B-active-parameter model advertised for Mandarin, English, Cantonese, dialectal, and quiet speech~\cite{glm2025modelcard}.

Consider a sentence in our pilot where Qwen3-ASR-1.7B transcribes \texttt{Claude Code} as \texttt{Cloud Code} at just 1.69\% sentence WER (Table~\ref{tab:examples}). The sequence metric records one word substitution, yet the complete product name is lost. Whole-phrase evaluation makes that failure explicit.

Low sequence error rates on some established ASR benchmarks coexist with continued production investment in ASR engineering. Some of that engineering addresses formatting and acoustic structure: voice activity detection, segmentation, timestamps, punctuation restoration, overlap handling, and speaker attribution. A separate class directly protects information-bearing terms by using contacts, meeting materials, domain lexicons, application state, or user vocabularies. Research and production-oriented systems repeatedly add contextual biasing, hotword retrieval, phonetic entity recovery, $n$-best reranking, and LLM post-correction for this second problem~\cite{li2024cbwhisper,asano2025contextual,ma2024asrec,recast2025,serrino2019contextual,huang2024neuralbiasing,xie2026nim4asr}.

We argue that this tension is partly an evaluation mismatch. Conventional metrics assign similar edit cost to a disposable filler and a product name whose corruption makes a transcript unusable. Prior ASR studies have therefore proposed keyword-, entity-, task-, and semantic-aware evaluation, and have shown that these measures correlate differently with downstream utility and human judgments than raw WER~\cite{park2008ker,galibert2016task,rugayan2023asd}.

This paper makes three contributions:
\begin{enumerate}
    \item We define whole-phrase preservation as an ASR evaluation task covering sample-local entities, technical terms, versioned names, and relational phrases.
    \item We operationalize prior KER ideas in a reproducible bilingual protocol with frozen annotations, shared reference/output normalization, exact-span validation, and scorer invariants.
    \item We evaluate seven systems on a 10-hour model-conditioned challenge set and a disjoint 30-minute model-blind control, documenting phrase errors across both sets, language-specific differences, and failures within low-sequence-error clips.
\end{enumerate}

\begin{figure*}[t]
\centering
\includegraphics[width=\textwidth]{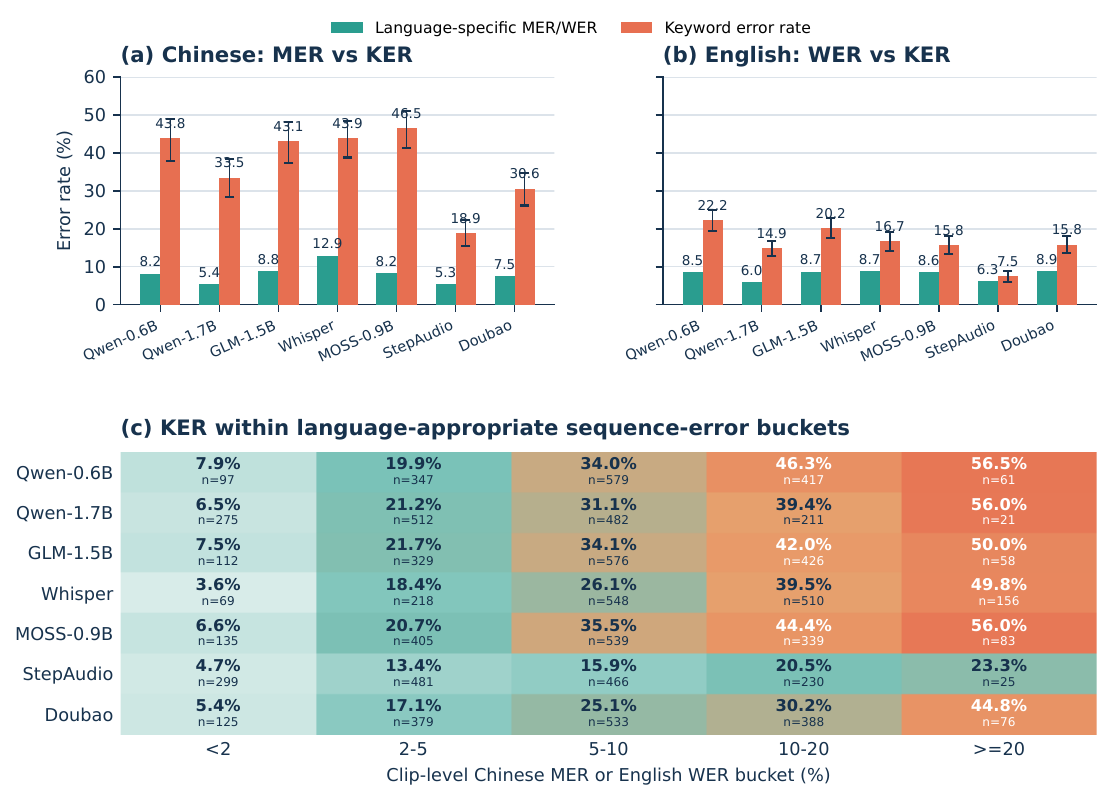}
\caption{Sequence error and phrase-occurrence error across seven frozen system configurations. Panels (a) and (b) show KER beside the language-appropriate conventional metric; error bars are 95\% recording-group bootstrap intervals for KER. Panel (c) groups clips into each system's own sequence-error buckets. Each cell reports KER and clip count.}
\label{fig:mismatch}
\end{figure*}

\section{Background and Related Work}

\subsection{Real-World Speech Corpora}

Public corpora increasingly include realistic conditions. GigaSpeech provides 10,000 hours of multi-domain English speech~\cite{chen2021gigaspeech}. WenetSpeech collects web video and podcast audio and includes a real-meeting test set~\cite{zhang2022wenetspeech}. SPGISpeech contains 5,000 hours of professionally transcribed earnings calls~\cite{oneill2021spgispeech}. These resources broaden acoustic and domain coverage; phrase-level annotations add an explicit account of the content to preserve.

Earnings calls make the role of context especially explicit. ConEC augments Earnings-21/22 with slides, earnings releases, participant names, and affiliations for contextual ASR~\cite{huang2024conec}. Contextual Earnings-22 targets rare, context-defined custom vocabulary and reports both no-context and supplied-context conditions; a keyword true positive requires the text and its aligned position to be correct~\cite{durmus2026contextualearnings22}. It is the closest existing benchmark to our motivation, studying contextual interventions in an English domain. Company names, product lines, executives, and values are precisely the content needed by retrieval, summarization, and decision systems. Similar issues occur in courses, where terminology and formula names dominate utility, and in podcasts, where guest names and code-switched product terms are long-tailed.

\subsection{Beyond Uniform WER}

WER counts substitutions, deletions, and insertions uniformly after sequence alignment; CER applies the same principle at character level. Galibert et al. showed that ASR error importance for spoken named entity recognition is ranked differently by task-relevant metrics and raw WER frequency~\cite{galibert2016task}. Semantic and syntactic metrics such as AraDiaWER combine multiple scores through learned or selected weights~\cite{ali2023aradiawer}.

Weighted metrics expose a practical ambiguity. Even if an entity is more important than a filler, it is unclear whether its cost should be 2, 5, or 10 across subtitles, meetings, finance, and in-vehicle control. Keyword selection expresses importance through binary inclusion, with phrase boundaries and aliases fixed by annotation. Park et al. studied the relation between WER and KER in contact-center analytics and defined KER over domain keywords~\cite{park2008ker}. VietMed-NER also reports KER, but uses alignment and excludes insertions; our profile deliberately discards position and penalizes only extra occurrences of selected keywords~\cite{leduc2025medicalner}. Building on this metric family, we contribute a bilingual whole-phrase protocol and a two-set challenge/control study.

Task-aware evidence extends beyond keyword counting. Galibert et al. compare WER with named-entity WER and a downstream-impact metric, finding that error importance is ranked differently by the chosen objective~\cite{galibert2016task}. Rugayan et al. show that Aligned Semantic Distance correlates better than WER with human judgments and with downstream named-entity and sentiment performance~\cite{rugayan2023asd}. NaturalCodeBench identifies a related mismatch between saturated canonical benchmarks and real-world utility for natural coding requests~\cite{zhang2024naturalcodebench}.

\subsection{Contextual Biasing and Correction}

Long-tail entities remain difficult for domain-general recognizers. Contextual recovery of out-of-lattice entities reduced lattice WER from 11.5\% to 4.9\% on an entity test set by combining phonetic similarity and context~\cite{serrino2019contextual}. CB-Whisper detects open-vocabulary keywords before decoding and injects spoken-form hints~\cite{li2024cbwhisper}. RECAST retrieves candidates from large dictionaries and prompts a downstream speech model, reporting large relative entity-WER improvements~\cite{recast2025}.

Post-ASR correction is complementary. Ma et al. compare 1-best, $n$-best, lattice-constrained, cross-recognizer, and zero-shot LLM correction~\cite{ma2024asrec}. Asano et al. use dialogue state and task context to rank candidate corrections in goal-oriented dialogue~\cite{asano2025contextual}. Their results further motivate measuring the keyword failures that make correction necessary.

Industrial work provides the same signal. Google describes song, app, and contact-name biasing as a production requirement for end-to-end ASR~\cite{zhao2019shallow}. Apple retrieves phonetically similar names from personal databases and reports a 73.6\% relative named-entity error reduction on a voice-assistant task~\cite{lei2024phonetic}. A Xiaomi-affiliated study injects contacts or user vocabulary into contextual ASR~\cite{huang2024neuralbiasing}; NIO's NIM4-ASR combines streaming recognition with retrieval-augmented, million-scale hotword customization~\cite{xie2026nim4asr}; and contact-center work uses LLM-generated anchors to repair phonetically similar brand and business names after transcription~\cite{koilakuntla2024contact}. Together, these interventions motivate evaluating critical-phrase preservation alongside end-to-end WER.

\section{Task Definition}

\subsection{Whole-Phrase Keywords}

Each selected utterance contains one or more sample-local keyword phrases. A keyword is a complete span whose corruption changes the useful information of that utterance:
\begin{itemize}
    \item entities such as people, organizations, products, and locations;
    \item domain terminology from technical, educational, or financial contexts;
    \item versioned phrases such as \texttt{Kimi 2.7} or \texttt{GPT-4o}, rather than every number in isolation;
    \item relational phrases whose partial deletion changes a named referent, relation, or requested action;
    \item other exact spans selected as necessary for the declared task.
\end{itemize}

Keywords are contextual. \texttt{Claude Code} can be the required phrase in a comparison of coding agents, while \texttt{Claude} may be an approved variant when only the vendor family matters. Equivalence is local to a sample and is frozen before model evaluation.

\subsection{Keyword Error Rate}

Reference and hypothesis are normalized with Unicode NFKC and case folding. Keyword spelling ignores layout spaces, underscores, and hyphens; other punctuation, including decimal points, slashes, and ampersands, is preserved. Numeric/version boundaries prevent \texttt{2.7} from matching \texttt{1.2.7} or \texttt{2.7.1}, and prevent \texttt{GPT-4} from matching \texttt{GPT-4.1}. Numerically equivalent notation or spoken forms must be frozen as sample-local variants. For each selected keyword $k$, let $N_k$ be its reference occurrence count and $C_k$ its non-overlapping normalized occurrence count in the hypothesis. Missing occurrences are $M_k=\max(0,N_k-C_k)$ and extra selected-keyword occurrences are $I_k=\max(0,C_k-N_k)$. Micro-averaged KER is
\begin{equation}
    \mathrm{KER}=\frac{\sum_k (M_k+I_k)}{\sum_k N_k}.
\end{equation}
Repeated occurrences are counted separately. Ordinary words, fillers, formatting, and non-keyword insertions do not enter the numerator or denominator. KER can exceed 100\% when selected-keyword insertions outnumber reference occurrences. Reference and output counts use the same production matcher; regression tests require reference-to-reference KER to equal zero.

KER measures preservation of selected phrase counts, with inclusion and phrase boundaries specified by frozen annotations. Systems produce full transcripts, and we report MER/WER alongside KER to capture sequence-wide errors.

\subsection{Annotation Protocol}

We use a versioned, auditable text-review pipeline:
\begin{enumerate}
    \item start from the benchmark reference transcripts;
    \item seed exact-span candidates with entity and terminology rules, matching longer phrases before subphrases;
    \item ask Codex CLI to retain one to three information-bearing names, products, technical terms, versions, or relational phrases, while rejecting ordinary words, program titles, and boilerplate;
    \item validate mechanically that every retained phrase is an exact normalized reference span and that every clip has at least one keyword;
    \item manually review and spot-check cleaned references, retained clips, and keyword annotations, then freeze keywords, variants, source groups, internal audio fingerprints, and scorer version.
\end{enumerate}

Every retained keyword must pass exact-span validation against its reference. Aliases are admitted only when an alternate written form is explicitly present in the same reference during this pilot.

\section{Pilot Definition and Versioning}

The paper reports pilot release v0.1. The internal directory name \texttt{hard10\_v2} denotes the second construction pass, while the repaired private manifest schema is v0.3 after unifying reference and hypothesis counting. Every row now declares that schema. A migration audit reconstructs and verifies the legacy inference-manifest hash, records the 44 challenge and two control denominator changes, asserts that protected audio/reference/keyword fields are unchanged, and binds the migrated manifest to prediction and scorer hashes. Historical inference receipts retain their original hashes rather than being rewritten. These identifiers describe different layers and do not denote different audio sets.

\subsection{Scope and Sources}

The current 10-hour pilot primarily contains Chinese and English podcasts, roundtable conversations, and technical discussions. These long-form sources naturally contain disfluency, self-repair, code-switching, long-tail entities, and cross-segment context.

\subsection{Cleaning, Selection, and Leakage Control}

We split by complete recording and remove non-speech, music, program edges, intros/outros, sponsor reads, and low-confidence references. Source caps and internal fingerprints are frozen in a non-public registry; detailed thresholds are recorded in a versioned internal protocol.

The pilot is a model-conditioned challenge set. Three systems---Qwen3-ASR-0.6B, Qwen3-ASR-1.7B, and GLM-ASR-Nano-2512---were pre-specified before final evaluation. An upstream, versioned selection diagnostic marked whether each system met or exceeded 5\% sequence error. Its Chinese unit was a compact character error that retained Latin characters and applied numeric normalization; English used word error. Within each language/source-origin cell, the construction targeted 55\%, 25\%, and 20\% of duration from clips marked wrong by at least two, exactly one, or none of the three systems, then ranked eligible clips by keyword coverage, label confidence, and the frozen diagnostic. The systems need not make the same lexical error; Whisper, MOSS, StepAudio, and Doubao were not used. The historical diagnostic labels 55.0\% of selected clips at $\geq2/3$, whereas a post-hoc audit using the report tokenizer and a $>5\%$ threshold gives 64.6\% and changes 555 clip tiers. We retain the historical labels to describe what actually drove sampling.

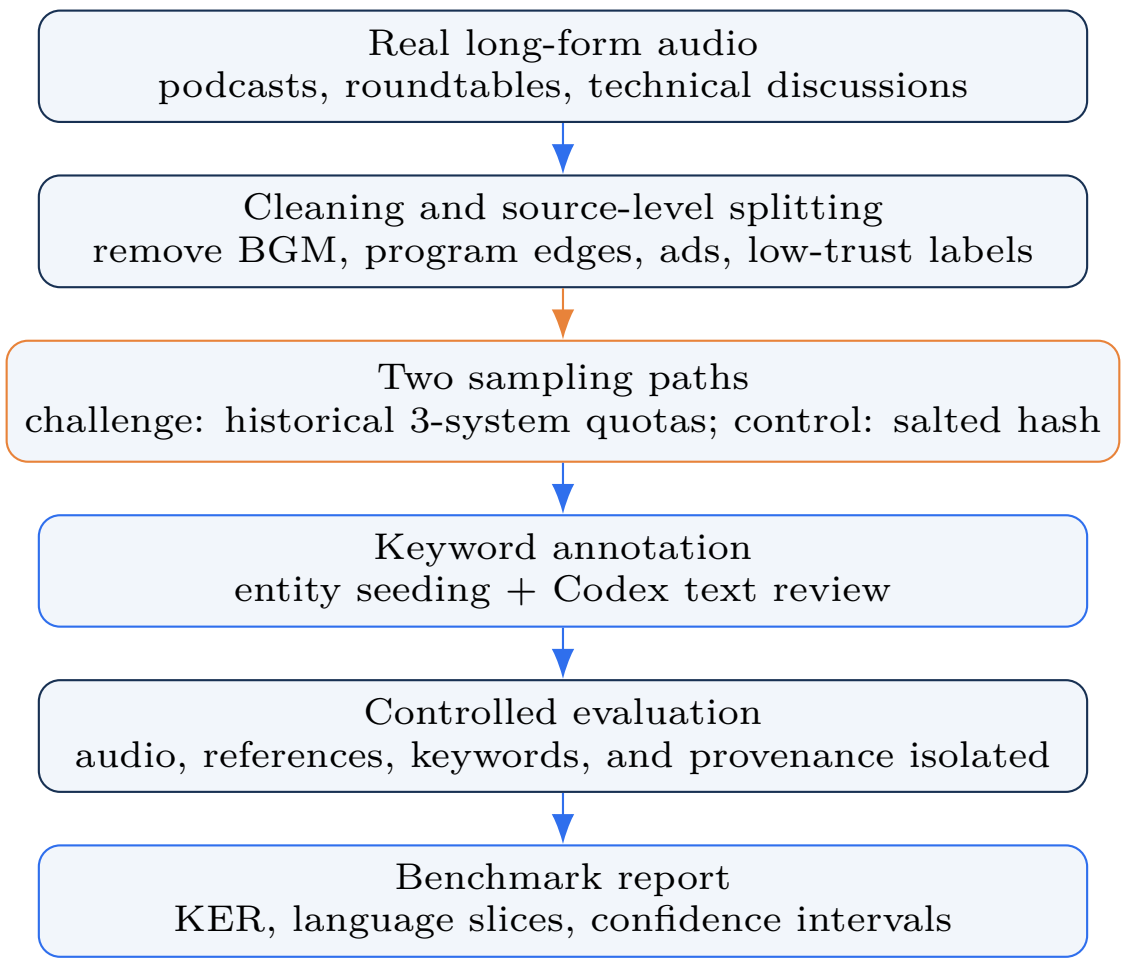
\begin{figure}[t]
\centering
\begin{tikzpicture}[node distance=3.4mm, every node/.style={font=\scriptsize,align=center}]
\node[draw=navy,rounded corners,fill=light,minimum width=0.83\columnwidth,minimum height=6mm] (a) {Real long-form audio\\podcasts, roundtables, technical discussions};
\node[draw=navy,rounded corners,fill=light,below=of a,minimum width=0.83\columnwidth,minimum height=6mm] (b) {Cleaning and source-level splitting\\remove BGM, program edges, ads, low-trust labels};
\node[draw=orange,rounded corners,fill=light,below=of b,minimum width=0.83\columnwidth,minimum height=8mm] (s) {Two sampling paths\\challenge: historical 3-system quotas; control: salted hash};
\node[draw=blue,rounded corners,fill=light,below=of s,minimum width=0.83\columnwidth,minimum height=6mm] (c) {Keyword annotation\\entity seeding + Codex text review};
\node[draw=navy,rounded corners,fill=light,below=of c,minimum width=0.83\columnwidth,minimum height=6mm] (d) {Controlled evaluation\\audio, references, keywords, and provenance isolated};
\node[draw=blue,rounded corners,fill=light,below=of d,minimum width=0.83\columnwidth,minimum height=6mm] (e) {Benchmark report\\KER, language slices, confidence intervals};
\draw[-{Latex[length=2mm]},blue] (a)--(b);
\draw[-{Latex[length=2mm]},orange] (b)--(s);
\draw[-{Latex[length=2mm]},blue] (s)--(c);
\draw[-{Latex[length=2mm]},blue] (c)--(d);
\draw[-{Latex[length=2mm]},blue] (d)--(e);
\end{tikzpicture}
\caption{Construction and evaluation workflow.}
\label{fig:pipeline}
\end{figure}

\subsection{Statistics and Reporting Policy}

\begin{table}[t]
\centering
\caption{KeyASRBench v0.1 pilot statistics.}
\label{tab:stats}
\scriptsize
\begin{tabular}{@{}p{0.53\columnwidth}>{\raggedleft\arraybackslash}p{0.39\columnwidth}@{}}
\toprule
Property & Value \\
\midrule
Total duration & 9.999 h \\
Chinese / English & 5.000 / 5.000 h \\
Segments & 1,501 (750 Chinese / 751 English) \\
Source-origin partitions A / B & 751 / 750 clips \\
Recording groups & 273 \\
Historical $\geq2/3$ selection label (quota target) & 55.0\% \\
Post-hoc $\geq2/3$ under report metric & 64.6\% \\
Selected keyword entries / occurrences & 3,881 / 4,723 \\
Keyword-bearing segments & 1,501 (100\%) \\
\bottomrule
\end{tabular}
\end{table}

\begin{figure*}[t]
\centering
\includegraphics[width=0.94\textwidth]{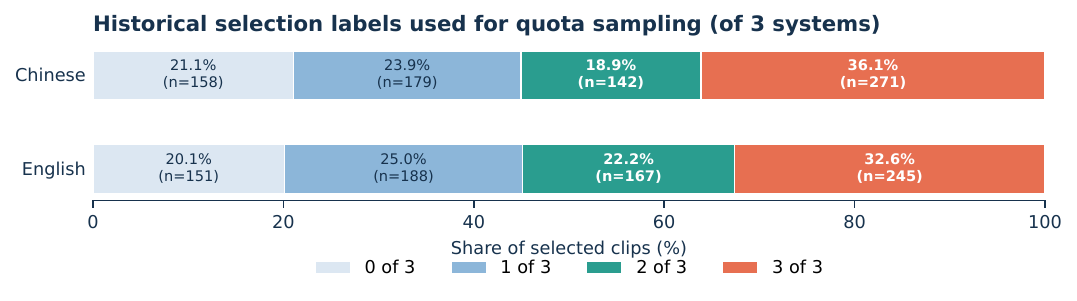}
\caption{Historical labels that drove quota sampling for the challenge set. They use the upstream selection-time normalizer, not the final report metric. The target composition was 55\%/25\%/20\% of duration from clips marked wrong by at least two, exactly one, or none of the three systems. Recomputing with the final report metric gives 64.6\% at $\geq2/3$ and is reported only as a post-hoc audit.}
\label{fig:selection-profile}
\end{figure*}

Every selected clip in the fixed 10-hour budget contains at least one keyword. The Chinese subset is defined by the clip's primary language and discourse context. Code-switched terms such as \texttt{AMD}, \texttt{DeepSeek}, and \texttt{Claude Code} remain part of Chinese clips when they are the terms actually spoken. Every release reports both keyword coverage and denominator size.

The 3,881 entries are sample-local annotations rather than globally unique lexical types. Dominant categories are product/model (1,530 entries), organization (1,020), and domain term (758); relational phrases (88) and versions/identifiers (29) are exploratory. Thirty-eight challenge entries and two control entries contain an approved alias under the narrow same-reference policy.

The planned public package contains the technical report, protocol, scorer, aggregate statistics, and separately cleared examples, not the complete benchmark data. Audio, references, keyword labels, and detailed provenance remain in a controlled environment.

\subsection{Data Boundary and Governance}

The report withholds program names, publishers, speakers, titles, URLs, timestamps, source IDs, recording-level durations, show notes, and audio hashes that could support reverse lookup. Any example audio released with the report must be separately licensed or explicitly cleared and must not overlap the controlled evaluation recordings.

Raw audio and source metadata remain access-controlled with internal fingerprints and provenance records. The current manifests are approved for this internal pilot but still carry a release-clearance requirement. Any future public examples, third-party evaluation access, retention policy, and deletion process require separate documentation and clearance.

\section{Experiments}

\subsection{Models and Input Contract}

Five systems are inferred locally from frozen checkpoints: Qwen3-ASR-0.6B, Qwen3-ASR-1.7B, GLM-ASR-Nano-2512 (1.5B active parameters), Whisper large-v3-turbo, and MOSS-Transcribe-Diarize-0.9B. The three selection systems use greedy decoding with a 512-token cap and no external context. Whisper uses faster-whisper 1.2.1 with CTranslate2 4.8.4 in FP16, beam size 1, temperature 0, forced manifest language, no VAD, and no previous-text conditioning. MOSS uses deterministic decoding in BF16 with a 1,024-token cap and language-specific neutral transcription prompts; speaker tags are removed mechanically before scoring. Model files, backend versions, prompts, manifest hashes, and decoding receipts are frozen internally; immutable upstream revisions unavailable from the local snapshots are marked unavailable rather than inferred. Whisper large-v3-turbo is the 809M-parameter pruned, faster variant of large-v3~\cite{openai2024whisperturbo}.

The other two systems use providers' official APIs. StepFun is called with model ID \texttt{stepaudio-3-asr-max}, mono 16-kHz PCM, forced language, ITN enabled, and timestamps disabled. Volcengine is called with resource ID \texttt{volc.seedasr.auc}, displayed as Doubao ASR 2.0, with the source audio format, forced language, ITN, punctuation, and utterance output enabled~\cite{volc2026seedasr}. Challenge requests completed on 2026-10-31 UTC and control requests on 2026-11-07 UTC. No system receives references, keywords, titles, hotwords, or retrieval results. Outputs, success receipts, and request hashes are frozen before scoring, identifying each API run by its dated configuration. The StepAudio technical report describes the model family, not an immutable revision of this endpoint~\cite{stepfun2026stepaudio3,bai2024seedasr}.

KER is the primary metric. Chinese mixed error rate (MER) and English WER provide language-specific sequence diagnostics, reported separately to retain their respective token units.

\subsection{Evaluation Protocol}

For both sequence diagnostics, NFKC/case-folded text is tokenized as one CJK character or an ASCII alphanumeric run that may retain internal \texttt{.}, \texttt{+}, \texttt{\#}, or \texttt{-}. Other punctuation delimits tokens; for example, \texttt{lock-in} and \texttt{a.b} are each one token, while an apostrophe splits \texttt{don't}. Chinese clips report this mixed tokenization as MER and English clips as WER. KER uses the separate whole-phrase procedure in Section~3.2 and accumulates missing and extra selected-keyword occurrences. Thus, corrupting the only keyword in a clip produces clip KER of 100\%, even when the rest of a long transcript is correct; emitting it twice produces one keyword insertion.

We micro-average KER across occurrences. Confidence intervals resample the 273 recording groups with replacement 10,000 times, preserving all clips and keyword occurrences within each sampled group; the reported interval is the percentile 95\% interval. This preserves within-recording dependence among the 4,723 occurrences and 1,501 clips. The intervals describe uncertainty in each system's score. Table~\ref{tab:examples} reports sentence-level diagnostics over illustrative case sentences.

\subsection{Aggregate Results}

\begin{table*}[t]
\centering
\caption{ASR-system results on the 10-hour challenge set. Parentheses give 95\% recording-group bootstrap intervals for overall KER. StepAudio and Doubao use dated official API configurations; the other five systems are inferred locally. No system receives a keyword list or external context.}
\label{tab:results}
\scriptsize
\setlength{\tabcolsep}{4pt}
\renewcommand{\arraystretch}{1.06}
\begin{tabular}{lccccc}
\toprule
Model & Chinese MER $\downarrow$ & English WER $\downarrow$ & Overall KER $\downarrow$ & Chinese KER $\downarrow$ & English KER $\downarrow$ \\
\midrule
Qwen3-ASR-0.6B & 8.19\% & 8.52\% & 33.81\% (29.38--38.14) & 43.81\% & 22.20\% \\
Qwen3-ASR-1.7B & 5.35\% & \textbf{5.95\%} & 24.88\% (21.12--28.58) & 33.49\% & 14.87\% \\
GLM-ASR-Nano-2512 & 8.77\% & 8.71\% & 32.50\% (28.11--36.67) & 43.07\% & 20.23\% \\
Whisper large-v3-turbo & 12.91\% & 8.74\% & 31.29\% (26.67--35.62) & 43.89\% & 16.66\% \\
MOSS-Transcribe-Diarize-0.9B & 8.22\% & 8.65\% & 32.29\% (27.09--36.93) & 46.49\% & 15.79\% \\
StepAudio 3 ASR Max & \textbf{5.31\%} & 6.27\% & \textbf{13.61\%} (10.90--16.27) & \textbf{18.87\%} & \textbf{7.51\%} \\
Doubao ASR 2.0 & 7.47\% & 8.88\% & 23.78\% (20.55--26.97) & 30.61\% & 15.84\% \\
\bottomrule
\end{tabular}
\renewcommand{\arraystretch}{1.0}
\end{table*}

Across the seven systems, language-appropriate conventional error ranges from 5.31\% to 12.91\%, while occurrence-based KER ranges from 13.61\% to 33.81\% (Table~\ref{tab:results}). StepAudio has the lowest point estimate for overall KER and Chinese MER; Qwen3-ASR-1.7B has the lowest English WER. Their language-specific rankings differ: StepAudio and Qwen3-ASR-1.7B differ by 0.04 points in Chinese MER but by 14.62 points in Chinese KER. In English, Qwen3-ASR-1.7B has a 0.32-point WER advantage while StepAudio has a 7.36-point KER advantage. Sequence accuracy and selected-phrase preservation thus order these systems differently.

Chinese KER is higher than English KER on the challenge set. Its keywords are dominated by technical entities, and Chinese clips contain many English product names, acronyms, and code-switched phrases.

Figure~\ref{fig:mismatch}(c) shows selected-phrase errors within low-sequence-error clips. Across the challenge set, the primary denominator contains 2,888 one-unit, 1,191 two-unit, and 644 three-or-more-unit occurrences. KER varies across these length groups, with patterns that differ across systems.

\subsection{Model-Blind Control and Descriptive Set Differences}

The 10-hour set prioritizes errors from three pre-specified systems and may therefore disadvantage them. We use a disjoint control sampled by a fixed salted hash after keyword-eligibility review, without reading model predictions, error rates, or difficulty scores. The control contains 77 clips, 61 recording groups, 0.504 hours of audio, and 216 keyword occurrences, with approximately 15 minutes per language and no recording-group overlap with the challenge set. It uses the same text-review policy, manual review/spot checks, exact-span validator, and corrected scorer. Intervals resample complete recording groups 10,000 times.

The control is model-blind but not source-matched. The challenge set contains 751/750 clips from the two internal source-origin partitions, whereas the control contains 76/1; source recordings, keyword mix, and density also differ. With 38 Chinese and 39 English clips (27 and 34 recording groups), the control provides an existence check with wide intervals. It does not support fine-grained ranking or natural-traffic estimation, and cross-set deltas cannot be attributed uniquely to model-conditioned selection.

\begin{table*}[t]
\centering
\caption{KER on the model-blind control and challenge set. Parentheses give 95\% recording-group bootstrap intervals for control KER.}
\label{tab:control}
\scriptsize
\setlength{\tabcolsep}{6pt}
\begin{tabular}{lccc}
\toprule
System & Model-blind control KER $\downarrow$ & 10-hour challenge KER $\downarrow$ & Challenge $-$ control \\
\midrule
Qwen3-ASR-0.6B & 24.54\% (17.79--31.78) & 33.81\% & +9.27 pp \\
Qwen3-ASR-1.7B & 15.74\% (10.30--21.49) & 24.88\% & +9.14 pp \\
GLM-ASR-Nano-2512 & 21.30\% (14.56--28.38) & 32.50\% & +11.20 pp \\
Whisper large-v3-turbo & 27.31\% (19.07--35.81) & 31.29\% & +3.98 pp \\
MOSS-Transcribe-Diarize-0.9B & 25.46\% (18.57--33.18) & 32.29\% & +6.83 pp \\
StepAudio 3 ASR Max & \textbf{10.65\%} (6.79--14.75) & \textbf{13.61\%} & +2.96 pp \\
Doubao ASR 2.0 & 21.30\% (15.02--27.71) & 23.78\% & +2.48 pp \\
\bottomrule
\end{tabular}
\end{table*}

All seven systems have 10.65--27.31\% KER on the model-blind control, establishing that selected-phrase errors also occur outside clips chosen from the three selection systems' outputs. The selection systems show challenge--control increases of 9.14--11.20 points, while the other four show 2.48--6.83 points (Table~\ref{tab:control}).

\subsection{Ground-Truth and Model-Output Examples}

\begin{table*}[t]
\centering
\caption{Real bilingual pilot examples. All keywords within a case occur in the same sentence. Ground truth gives Codex-selected exact reference spans; recognized forms are copied from frozen model outputs. MER/WER and KER are computed only on that case sentence. Repeated occurrences count separately. Red denotes a miss; black denotes a recovered phrase. Whisper is omitted from this illustrative table for space; aggregate results include all seven systems.}
\label{tab:examples}
\scriptsize
\setlength{\tabcolsep}{2pt}
\renewcommand{\arraystretch}{1.04}
\begin{tabular}{@{}lllp{0.34\textwidth}cc@{}}
\toprule
Case type & Ground truth & System & Recognized form(s) & Sent. MER/WER & Sent. KER \\
\midrule
\multirow{6}{*}{ZH-entity} & \multirow{6}{*}{\texttt{极客公园; 宇树科技}} & Qwen-0.6B & \zhmiss{极公源}; \zhhit{宇树科技} & 6.25\% MER & 50\% \\
 & & Qwen-1.7B & \zhmiss{极工源}; \zhmiss{语数科技} & 10.42\% MER & 100\% \\
 & & GLM-1.5B & \zhmiss{极光源}; \zhmiss{语数科技} & 10.42\% MER & 100\% \\
 & & MOSS-0.9B & \zhmiss{极光园}; \zhmiss{语数科技} & 8.33\% MER & 100\% \\
 & & StepAudio & \zhmiss{极公园}; \zhhit{宇树科技} & 10.42\% MER & 50\% \\
 & & Doubao & \zhmiss{极公园}; \zhhit{宇树科技} & 10.42\% MER & 50\% \\
\addlinespace[2pt]
\multirow{6}{*}{ZH-technical} & \multirow{6}{*}{\texttt{长程任务; 自治智能体}} & Qwen-0.6B & \zhmiss{长城任务}; \zhmiss{自制智能体} & 4.76\% MER & 100\% \\
 & & Qwen-1.7B & \zhmiss{长征任务}; \zhmiss{自制智能体} & 7.14\% MER & 100\% \\
 & & GLM-1.5B & \zhmiss{长征任务}; \zhmiss{自制智能体} & 9.52\% MER & 100\% \\
 & & MOSS-0.9B & \zhmiss{长征任务}; \zhmiss{自制智能体} & 11.90\% MER & 100\% \\
 & & StepAudio & \zhhit{长程任务}; \zhhit{自治智能体} & 4.76\% MER & 0\% \\
 & & Doubao & \zhmiss{长城任务}; \zhmiss{自制智能体} & 7.14\% MER & 100\% \\
\midrule
\multirow{6}{*}{EN-product} & \multirow{6}{*}{\texttt{Codex; Claude Code}} & Qwen-0.6B & \hit{Codex}; \miss{Cloud Code} & 5.08\% WER & 50\% \\
 & & Qwen-1.7B & \hit{Codex}; \miss{Cloud Code} & 1.69\% WER & 50\% \\
 & & GLM-1.5B & \miss{codec}; \miss{cloud code} & 8.47\% WER & 100\% \\
 & & MOSS-0.9B & \hit{Codex}; \hit{Claude Code} & 0.00\% WER & 0\% \\
 & & StepAudio & \hit{Codex}; \hit{Claude Code} & 1.69\% WER & 0\% \\
 & & Doubao & \hit{Codex}; \hit{Claude code} & 3.39\% WER & 0\% \\
\addlinespace[2pt]
\multirow{6}{*}{EN-entity} & \multirow{6}{*}{\texttt{Perplexity; HeyGen; Replit}} & Qwen-0.6B & \hit{Perplexity}; \miss{Hagen}; \hit{Replit} & 30.77\% WER & 33\% \\
 & & Qwen-1.7B & \hit{Perplexity}; \hit{HeyGen}; \miss{Replets} & 30.77\% WER & 33\% \\
 & & GLM-1.5B & \hit{Perplexity}; \miss{Hagen}; \miss{RepliX} & 46.15\% WER & 67\% \\
 & & MOSS-0.9B & \hit{Perplexity}; \miss{Hagen}; \miss{Replids} & 76.92\% WER & 67\% \\
 & & StepAudio & \hit{Perplexity}; \hit{HeyGen}; \miss{Ripples} & 69.23\% WER & 33\% \\
 & & Doubao & \hit{Perplexity}; \miss{Hagen}; \miss{Reblitz} & 23.08\% WER & 67\% \\
\bottomrule
\end{tabular}
\renewcommand{\arraystretch}{1.0}
\end{table*}

Table~\ref{tab:examples} presents four illustrative cases. The first Chinese case shows homophone-like substitutions in organization names. The second shows substitutions in technical phrases; StepAudio recovers both displayed phrases despite another character error in the sentence. The English product case turns \texttt{Claude Code} into \texttt{Cloud Code} for all three selection systems, while MOSS and both official API systems recover the complete product name. The English entity case contains three named products.

Every case is linked to its frozen manifest entry and predictions. Sentence text and metrics are extracted by prediction ID and scorer-token interval, including all recognized fillers within that interval. The exact-span validator applies the same normalization and counting rule to the examples and aggregate results.

The result is consistent with a point made explicitly in the Qwen3-ASR report: models that differ little on open benchmarks may exhibit meaningful quality gaps in real scenarios~\cite{shi2026qwen3asr}.

\section{Discussion}

\subsection{Scope of Whole-Phrase Evaluation}

The two metric families provide complementary views of a full transcript. KER tracks selected phrase counts, while MER/WER exposes errors across the sequence, including ordinary words. Reading them together distinguishes preservation of names and phrases from overall transcription accuracy. Readability, diarization, and semantic relations provide further dimensions for evaluation.

\subsection{Keyword Selection Is the Benchmark}

Keyword selection determines which information the benchmark asks a transcript to preserve. Frequency alone can miss rare entities and relational phrases, so the annotation policy considers their role in the utterance and favors complete spans. In v0.1, model outputs remain hidden from the Codex keyword reviewer, and sample-local phrases and aliases are frozen before scoring. Future domains may ground selection in participant lists, releases, product tables, slides, or glossaries, with each information source frozen and declared.

\subsection{Next Experiments}

The next study should use a larger source-matched control drawn by complete recording group and expand the currently sparse relational (88 entries) and version/identifier (29 entries) categories. Overlap, far-field, and accent slices would extend acoustic coverage. A second priority is to relate phrase preservation to retrieval performance and human transcript-utility judgments. Finally, evaluating the same system under audio-only, hotword, contextual-bias, and retrieval-assisted conditions would test which interventions recover the observed failures.

\section{Limitations, Ethics, and Governance}

The recording-group bootstrap for the challenge set and 77-clip control leaves higher-level series correlation unmodeled. API runs occurred seven days apart, and immutable server revisions are unavailable; service drift may contribute to cross-set differences.

Annotations combine Codex-assisted text selection with manual review and spot checks, rather than complete independent dual-annotator audio review; residual reference errors, false-positive keywords, and omitted aliases may remain. KER has not been calibrated against retrieval, task completion, or transcript-utility judgments. Its numerical gap from MER/WER also reflects different units and whole-phrase length.

Personal names, meeting identities, and sensitive attributes require de-identification and authorization checks before any external use.

\section{Conclusion}

All seven evaluated systems exhibit whole-phrase preservation failures on both the model-conditioned challenge set and the model-blind control. On the 10-hour bilingual challenge set, five local configurations and two official API configurations achieve language-specific conventional errors of 5.31--12.91\%, while KER is 13.61--33.81\%; on the 30-minute control, KER is 10.65--27.31\%. KeyASRBench makes these failures explicit through frozen whole-phrase annotations and occurrence-based evaluation alongside sequence metrics.

\section*{Reproducibility Statement}

The frozen challenge set contains 1,501 clips, 273 recording groups, and 4,723 keyword occurrences. The model-blind control contains 77 clips, 61 recording groups, and 216 occurrences, with zero recording-group overlap. Chinese and English each contribute approximately five hours to the challenge set and 15 minutes to the control. Qwen3-ASR-0.6B, Qwen3-ASR-1.7B, GLM-ASR-Nano-2512, Whisper large-v3-turbo, and MOSS-Transcribe-Diarize-0.9B are inferred locally; StepAudio 3 ASR Max (\texttt{stepaudio-3-asr-max}) and Doubao ASR 2.0 (\texttt{volc.seedasr.auc}) use official APIs. All systems receive audio and forced language only. Prediction IDs must exactly match each private manifest. Reference and hypothesis counts call one shared matcher, whose tests include numeric/version extension counterexamples and zero reference-to-reference KER. KER is micro-aggregated over repeated whole-phrase occurrences; extra selected-keyword occurrences are penalized, while non-keyword insertions are outside the metric. Row-level schemas, a legacy-to-current migration audit, and aggregate result provenance bind the manifests, predictions, matcher, and scorer hashes; historical inference identities remain unchanged and sensitive inputs remain access-controlled.

\footnotesize
\bibliographystyle{plain}
\bibliography{references}

\end{document}